%% file: pixel-wise-street-segmentation.tex
\documentclass[technote,a4paper,leqno]{IEEEtran}
\pdfoutput=1
\usepackage{amssymb, amsmath}

\usepackage{filecontents}

\RequirePackage{ifpdf}
\ifpdf \PassOptionsToPackage{pdfpagelabels}{hyperref} \fi
\RequirePackage{hyperref}
\usepackage{parskip}
\usepackage[pdftex,final]{graphicx}
\usepackage{csquotes}
\usepackage{bm}
\usepackage{blindtext}\usepackage{braket}
\usepackage{booktabs}
\usepackage{multirow}
\usepackage{pgfplots}
\pgfplotsset{compat=newest}
\usepackage{wasysym}
\usepackage[symbol]{footmisc}
\usepackage{float}
\usepackage{caption}
\usepackage{subcaption}
\makeatletter
\newcommand\mynobreakpar{\par\nobreak\@afterheading}
\makeatother
\usepackage[noadjust]{cite}

\usepackage[capitalise]{cleveref} 
\usepackage[binary-units,group-separator={,}]{siunitx}
\DeclareSIUnit\pixel{px}
\usepackage{glossaries}
\loadglsentries[main]{glossary}
\makeglossaries

\title{Pixel-wise Segmentation of Street with Neural Networks}
\author{%
\makebox[.4\linewidth]{Sebastian Bittel\thanks{\IEEEauthorrefmark{1} These authors contributed equally to this work}\IEEEauthorrefmark{1}}
\and \makebox[.4\linewidth]{Vitali Kaiser\IEEEauthorrefmark{1}}\\
\and \makebox[.4\linewidth]{Marvin Teichmann\IEEEauthorrefmark{1}} 
\and \makebox[.4\linewidth]{Martin Thoma\IEEEauthorrefmark{1}}} 

\hypersetup{
    pdfauthor   = {Sebastian Bittel, Vitali Kaiser, Marvin Teichmann, Martin Thoma},
    pdfkeywords = {recognition, machine learning, neural networks, classification, multilayer perceptron, deep learning},
    pdfsubject  = {Recognition},
    pdftitle    = {Pixel-wise Segmentation of Street with Neural Networks},
}

\crefname{table}{Table}{Tables}
\crefname{figure}{Figure}{Figures}

\usepackage{microtype}

\begin{document}
\maketitle
\input{abstract}

\input{ch1-introduction}
\input{ch2-related-work}
\input{ch5-frameworks}

\input{ch4-our-model}
\input{ch7-results}
\input{ch8-discussion}

\bibliographystyle{IEEEtranSA}
\bibliography{pixel-wise-street-segmentation}
\end{document}

%% file: glossary.tex
\newacronym{CNN}{CNN}{Convolutional Neural Network}
\newacronym{CUDA}{CUDA}{Compute Unified Device Architecture}
\newacronym{FCN}{FCN}{Fully Convolutional Network}
\newacronym{GPU}{GPU}{graphics processing unit}
\newacronym{MRF}{MRF}{Markov Random Field}
\newacronym{PCA}{PCA}{principal component analysis}
\newacronym{PyPI}{PyPI}{Python Package Index}
\newacronym{sst}{sst}{street segmentation toolkit}

%% file: abstract.tex
\begin{abstract}
Pixel-wise street segmentation of photographs taken from a drivers perspective
is important for self-driving cars and can also support other object
recognition tasks. A framework called SST was developed to examine the accuracy
and execution time of different neural networks. The best neural network
achieved an $F_1$-score of \SI{89.5}{\percent} with a simple feedforward
neural network which trained to solve a regression task.
\end{abstract}

%% file: ch1-introduction.tex

\section{Introduction}
\label{sec:introduction}
Pixel-wise segmentation of street is an important part of assisted and
autonomous driving~\cite{Tarel2009}. It can help to understand road scenes and
reduce the space to search for lane markings. Traditionally, road segmentation
is done with computer vision methods such as watershed
transformation~\cite{Beucher1990}. Recent advances in deep neural networks,
especially in computer vision, suggest that \glspl{CNN} might be able to
achieve higher classification accuracy on road segmentation tasks then those
traditional approaches.

This paper was written in the context of a machine learning hands-on course at
Research Center of Information Technology at the KIT (FZI), where self-driving
cars are developed. One requirement of self-driving cars is that the algorithms
need to be fast (e.g. classify an image in less than \SI{20}{\milli\second}).

\Cref{sec:related-work} mentions published work which influenced us in the
choice of our methods. The basics methods used are explained in
\cref{sec:concept}. It follows a description of the realization in
\cref{sec:realization} with a description of the used frameworks as well as
details about the developed SST framework. Models are explained in
\cref{sec:model} and evaluated in \cref{sec:evaluation}. Finally,
\cref{sec:discussion} summarizes the lessons we've learned and mentions how
our work can be continued.

%% file: ch2-related-work.tex

\section{Related Work}\label{sec:related-work}
Road segmentation is a subproblem of general scene parsing or segmentation. In
scene parsing every object in a scene is classified pixelwise with a label.
Whereas in road segmentation often only two classes exist and more assumptions
can be applied.\\
In the first publications, roads were usually annotated by color-based
histogram approaches and specific model knowledge. Examples are the in 1994
introduced approach~\cite{Beucher1990} using the watershed algorithm or
\cite{aly2008real} where roads were annotated indirectly by lane markings found
with a Hough transformation.\\
Later insights of general scene parsing where transferred and more generic
approaches like~\cite{6182716} have achieved remarkable results with a
\gls{MRF} and superpixels.\\ The impressive classification results of
\glspl{CNN} like AlexNet~\cite{krizhevsky2012imagenet} or
GoogLeNet~\cite{SzegedyLJSRAEVR14} during the Google ImageNet LSVRC-2010
contest, made \glspl{CNN} interesting for all kinds of computer vision problems
like segmentation.\\
With~\cite{long2014fully} Long and his team introduced a method for general
scene parsing based on \glspl{FCN} and a deconvolutional layer.\\
This approach is used as a blueprint to our implementation, described in
\cref{sec:model}. Therefore the main concepts are introduced in
\cref{sec:concept}.\\
Instead of creating a new model, they converted existing classification
\glspl{CNN} like AlexNet or GoogLeNet into \glspl{FCN}. The obtained heat maps for
every class were calculated for multiple resolutions and upscaled with
deconvolution layer interpolation to the original resolution. With a fully
connected convolutional layer in the end, the multiple outputs are combined
into one classification heat map for every class.\\

In~\cite{mohan2014deep} a approach is presented, which makes also use of a
\gls{CNN} in combination with deconvolution. In comparison to Long's network,
among others it is less deep and uses less convolutional then deconvolutional
layers. Furthermore the input image is divided in multiple patches and for each
patch a separate neural network was trained. Their model achieved the
best-recorded result on the same data set we use, which is described in
\cref{sec:datasets}.

\section{Concept}\label{sec:concept}
Two concepts are explained in this section: \Glspl{CNN} and \glspl{FCN}, which
were introduced in~\cite{long2014fully}.

\subsection{CNN}
A \gls{CNN} is a feedforward neural network with at least one convolutional
layer. In general, applying a function repeatedly over the output of other
functions is defined as a convolution. In context of \gls{CNN} a convolution
layer is a learned filter which is applied at all possible offsets over an
input image. The output of the filter can be seen as an abstraction of the
input. Convolutional layers are often combined with pooling layers, which
extract one single value out of a larger region. Examples for pooling layer
functions are the \verb+average+ or \verb+max+ function.\\
\gls{CNN}s can be used for classification with a fully connected layer at the
end. The upper part of \cref{fig:fcn} shows such a network
for classification. In the example the input image is classified as
\enquote{tabby cat}, because this class has the highest probability
(illustrated as bar chart).

\subsection{FCN}
\begin{figure}[htb]
    \centering
    \includegraphics[width=9cm]{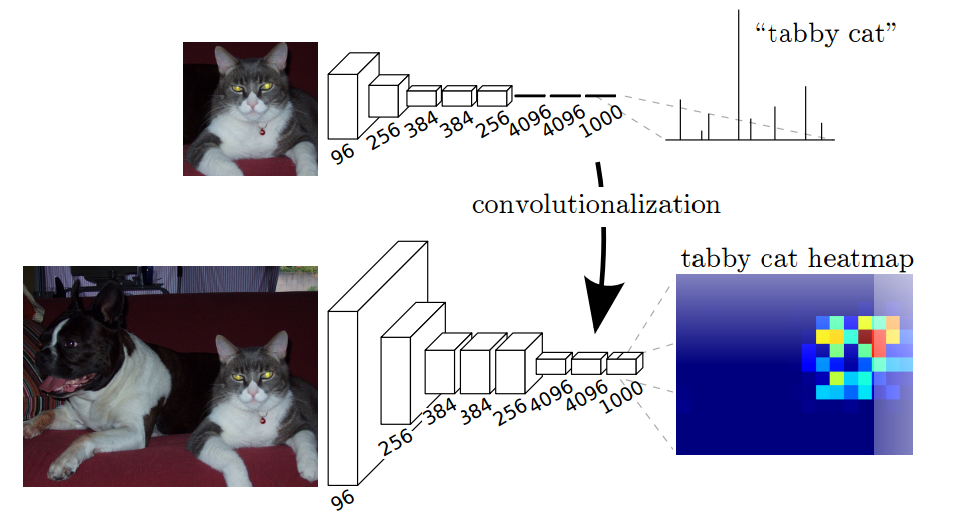}
    \caption{Comparison of a \gls{CNN} for classification (top) and a \gls{FCN} which creates a heat map (bottom). The image was taken from~\cite{long2014fully}.}%
\label{fig:fcn}
\end{figure}

A \gls{CNN} which consists only of convolutional or pooling layers is also
known as a \gls{FCN}. By removing fully connected nodes, the network can
generate a output for an image with arbitrary size. In~\cite{long2014fully}
this characteristic is used to train a network that produces a classification
heat map. In the lower part of \cref{fig:fcn}, a heat map of probabilities for
the presence of a \enquote{tabby cat} at different image regions is generated
from the input image.

\subsection{Fully connected convolutional layer}
A fully connected convolutional layer is a regular convolutional layer in size
of the input. Consequentially, the weight matrix covers every input neuron. Long
noted in~\cite{long2014fully} that it is the two-dimensional equivalent to a
fully connected layer in a classification \gls{CNN}.

\subsection{Deconvolutional Layer}
Deconvolutions are inverse convolutions. In context of neural networks the
function for forward and backward calculation are just switched. It is often
combined with unpooling and allows to learn how to reconstruct a larger image
region from a smaller input. Long indicates in~\cite{long2014fully} its
similarity to interpolation. He explains it as a highly parametrized non-linear
interpolation.

%% file: ch5-frameworks.tex

\section{Realization}\label{sec:realization}
\subsection{Existing Frameworks}\label{sec:frameworks}
nolearn was used in combination with Lasagne~\cite{sander_dieleman_2015_27878}
to train the models. Lasagne is based on Theano~\cite{Bergstra2010}. We also
used Caffe~\cite{Jia2014} for \glspl{CNN}, but skipped that approach as the
framework crashed very often for different training approaches without giving
meaningful error messages.

Theano is a Python package which allows symbolic computation of derivatives as
well as automatic generation of GPU code. This is used to calculate the weight
update function for arbitrary feed-forward networks. Lasagne makes using Theano
simpler by providing basic layer types like fully connected layers,
convolutional layers and pooling layers with their update function. nolearn
adds syntactic sugar and neural network objects with a similar interface
as the \verb+scikit-learn+ package uses for its classifiers~\cite{scikit-learn}.

\input{ch5-sst}

%% file: ch5-sst.tex

\subsection{SST}\label{sec:sst}
\Gls{sst} is a Python package hosted on \gls{PyPI} and developed on GitHub.
It makes use of Lasagne (see \cref{sec:frameworks}). It was mainly developed
during the course \enquote{Machine Learning Laboratory --- Applications} at
KIT by Sebastian Bittel, Vitali Kaiser, Marvin Teichmann and Martin Thoma.

\subsubsection{Installation}
\verb+sst+ can be installed via \verb+pip install sst+. To make the
installation as simple as possible, this does not try to install all
requirements.
\verb+sst selfcheck+ gives the user the possibility to check which packages
are still required and manually install them.

\subsubsection{Functionality}
\verb+sst+ makes use of Python files for neural network definitions. Those
models must have a \verb+generate_nnet(feature_vectors)+ function which
returns an object with a \verb+fit(features, labels)+ method, a
\verb+predict(feature_vector)+ method and a
\verb+predict_proba(feature_vector)+ method. This is typically achieved by
returning a \verb+nolearn+ model object. The Python network definition file
has to have two global variables: \verb+patch_size+ (a positive integer) and
\verb+fully+ (a Boolean). The first variable is the patch size expected by the
neural network, the second variable indicates if the neural network was trained
to classify each pixel of the complete patch (\verb+fully = True+) or only the
center pixel (\verb+fully = False+).

\verb+sst --help+ shows all subcommands. The subcommands are currently

\begin{itemize}
    \item \verb+selfcheck+: Test which components or Python packages are
                            missing and have to be installed to be able to use
                            all features of \verb+sst+.
    \item \verb+train+: Train a neural network.
    \item \verb+eval+: Evaluate a trained network on a photograph and also
                       generate an overlay image of the segmentation and the
                       data photograph.
    \item \verb+serve+: Start a web server which lets the user choose images
                        from the local file system to predict the label and to
                        show overlays.
    \item \verb+view+: Show all information about an existing model.
    \item \verb+video+: Generate a video.
\end{itemize}

%% file: ch4-our-model.tex

\section{Used models}\label{sec:model}

We have implemented two approaches to tackle  the street segmentation problem.
We have used a sliding window approach which is based on a classification
network and a regression approach. Both models are detailed in the following
section.

\subsection{The Sliding Window Approach}
Traditionally neural networks are used for classification tasks. As
described in \cref{sec:related-work} they deliver impressive results. Our first
approach is a sliding window model, which exploits the classification strength
of deep neural networks.

\subsubsection{Definition of the Classification Problem}
We trained a neural network to solve the following binary classification
problem:

\fbox{
    \begin{tabular}{l l}
        \multicolumn{2}{l}{\textbf{Classification Problem}} \\
        \textit{Input:} & A $n \times n$ 3-channel pixel image section.\\
        \textit{Output:} & Decide whether the center pixel is street.
    \end{tabular}
}

\begin{figure}[H]
    \centering
    \includegraphics[width=0.5\columnwidth]{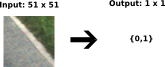}
    \caption{Visualization of the classification problem solved by our neural network.}%
\label{fig:figure}
\end{figure}

For $n$ we used $51$. This constant was chosen as we ran into GPU memory
problems when training on higher values of $n$. Our classification approach is
visualized in \cref{fig:figure}.

\subsubsection{Net Topology}
The problem defined above can be tackled with any of the well known
classification networks such as GoogLeNet or AlexNet. For our solution we
designed our own network detailed in \cref{tab:topo}. The small size of only
three hidden layers was chosen as our experiments showed that small networks
perform better than networks with more layers. One reason for this is that the
amount of labeled images is rather small. Hence smaller nets generalize much
better. Secondly, the binary decision task of recognizing street is much
simpler than detailed image classification. This simplicity does also reflect
in the net topology.

\begin{table}[H]
    \normalsize
    \centering
\begin{tabular}{r  l l}
    \toprule
    \textbf{Layer} & \textbf{Type}  & \textbf{Shape}  \\
    \midrule
    0     & Input &  $51 \times 51 \times 3$ \\
    1     & Convolution & 10 filter  each $5 \times 5$ \\
    2     & Convolution & 10 filter  each $5 \times 5$  \\
    3     & Pooling     & $2 \times 2$ \\
    4     & Output     & $1$ \\
    \bottomrule
\end{tabular}
\caption{Topology of the classification network.}
\label{tab:topo}
\end{table}

\subsubsection{Training}
The training data for this classification problem can be easily obtained by
modifying the original training data. One advantage of our approach is that we
get a lot of training data out of each image. In theory, we get one (distinct)
datum for each pixel in each training image. However, it is not useful to
actually use all of this data as patches which are close to each other and thus
are very similar. Hence the information gain of including these patches is very
small. On the other hand, if we generate an image section for each pixel we
obtain more data than the memory of our GPU can handle. We therefore introduced
a training stride. A stride of $s$ results in the center pixels having a
distance of $s$ in height and with to the next sections center pixel. This is
important in the section generation step before training as well as for the
pixel-wise classification. The overlap of two adjacent images is hence reduced
to $n-s$, where $n \times n$ is the size of each image section. Empirical
evaluations indicated that $s=10$ is a good default value for the trainings
stride.

\subsubsection{Evaluation}
\begin{figure}[H]
    \centering
    \includegraphics[width=\columnwidth]{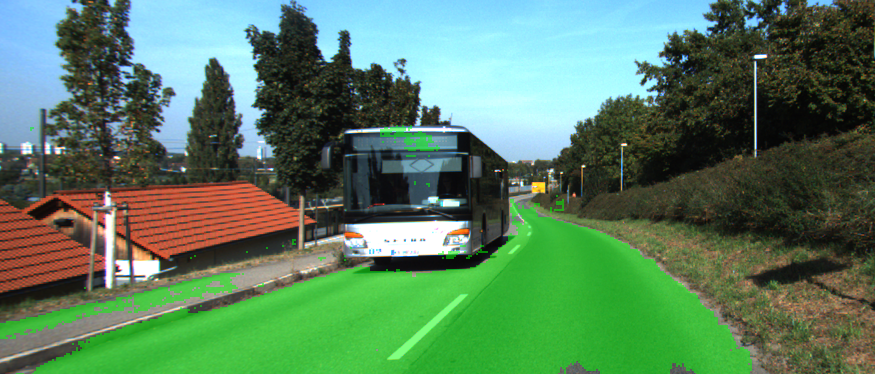}
    \caption{Using the sliding window approach with a stride of $s=2$.}%
\label{fig:stride2}
\end{figure}

In order to apply a classification network on the segmentation problem we used
the well known sliding window approach. The main idea is to apply the
classification network on each pixel $p$ of the input image by generating the
$n \times n$ image section with center pixel $p$. We use padding to be able to
apply the method to pixels close to the border. \Cref{fig:stride2} shows the
result of this approach.

\begin{figure}[H]
    \centering
    \includegraphics[width=\columnwidth]{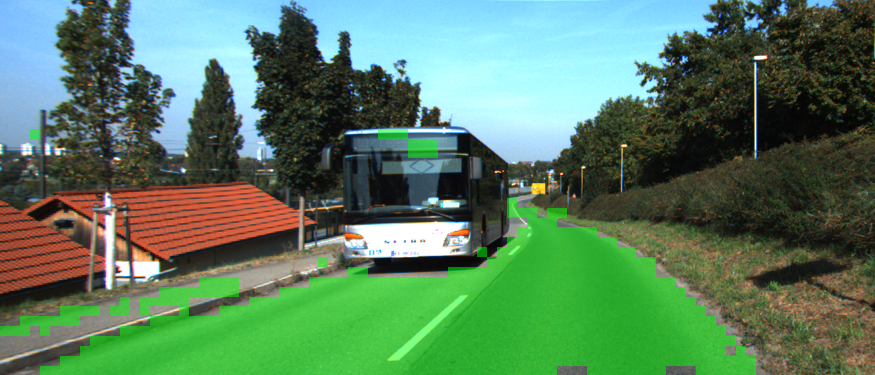}
    \caption{Using the sliding window approach with a stride of $s=10$.}%
\label{fig:stride10}
\end{figure}

The main disadvantage of this approach is the impractical runtime. For a
1~megapixel image we need to run 1~million classifications. This leads to a
runtime of almost two~minutes with our hardware. In order to reduce the
evaluation time we introduced an evaluation stride $s$. Similar to the training
stride we skip $s-1$ pixels in each dimension. This increases the evaluation
speed by a factor of $s^2$. For the sliding window approach we found that a
stride of $s = 10$ is a reasonable trade-off between speed and quality.
\Cref{fig:stride10} shows the result of the sliding window approach with a
stride of $s=10$.

\subsection{The Regression Approach}
The main disadvantage of our sliding window approach is that the segmentation
becomes very coarse with higher values for the stride $s$. To overcome this
problem we designed a regression neural networks which is able to classify each
pixel independently.

\subsubsection{Definition of the Regression Problem}
\begin{figure}[H]
    \centering
    \includegraphics[width=0.5\columnwidth]{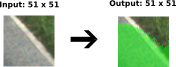}
    \caption{Visualization of the regression approach.}%
\label{fig:reg}
\end{figure}

We trained a neural network to solve the following regression problem:

\fbox{
    \begin{tabular}{l l}
        \multicolumn{2}{l}{\textbf{Regression Problem}} \\
        \textit{Input:} & A $n \times n$ 3-channel pixel image section.\\
        \textit{Output:} & A $n \times n$ label.
    \end{tabular}
}

where the net is trained to minimize the mean squared error of the output. The
output of the regression net is continuous. We round the output to obtain a
binary classification. \Cref{fig:reg} visualizes our regression approach.

The goal is to choose $n$ as big as possible as $n^2$ is the number of pixels
which can be classified at once. However, due to GPU memory limitations we
cannot train a network with $n > 51$.

\subsubsection{Net Topology}
Similar to the classification approach our experiments show that a simple net
topologies work best. The topology we used is detailed in \cref{tab:topo2}.

\begin{table}[H]
    \normalsize
    \centering
    \begin{tabular}{r l l}
        \toprule
        \textbf{Layer} & \textbf{Type}  & \textbf{Shape}  \\
        \midrule
        0     & Input &  $51 \times 51 \times 3$ \\
        1     & Convolution & 10 filter  each $5 \times 5$ \\
        2     & Convolution & 1 filter $51 \times 51$  \\
        3     & Reshape (Flatten) & $51 \times 51$ \\
        4     & Output     & $2601\footnotemark \times 1$\\
        \bottomrule
    \end{tabular}
    \caption{Topology of the regression network.}%
\label{tab:topo2}
\end{table}
\footnotetext{The shape of $2601 \times 1$ is a result of flattening the $51 \times 51$ image patch. This is only necessary due to tooling support.}

\subsubsection{Training}
Training of the regression model can be implemented analogously to the
classification model. We use overlapping image section again to get as much
information out of the data as possible.

\subsubsection{Evaluation}
\begin{figure}[tbp]
    \centering
    \includegraphics[width=\columnwidth]{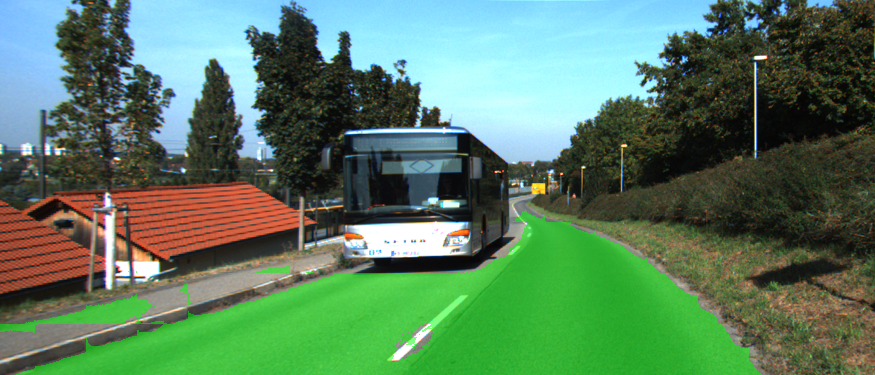}
    \caption{Using regression approach with stride $s=51$}%
\label{fig:reg_stride2}
\end{figure}

In order to evaluate a whole image using the regression approach we divide the
image into patches of size $n \times n$ and evaluate each patch individually.
The output is shown in \cref{fig:reg_stride2}. The result is quite impressive,
especially regarding the overall runtime of about $\SI{0.18}{\second}$ as shown
in \cref{tab:runtime}.

One observation is, that the segmentation works better in the center of each
patch. The neural network does not have good information close to the border of
each image section. To overcome this problem we use an evaluation stride again.
This introduces an overlap between each image patch. A pixel $p$ is then
segmented according to the patch whose center is closer to $p$.

\begin{figure}[tbp]
    \centering
    \includegraphics[width=\columnwidth]{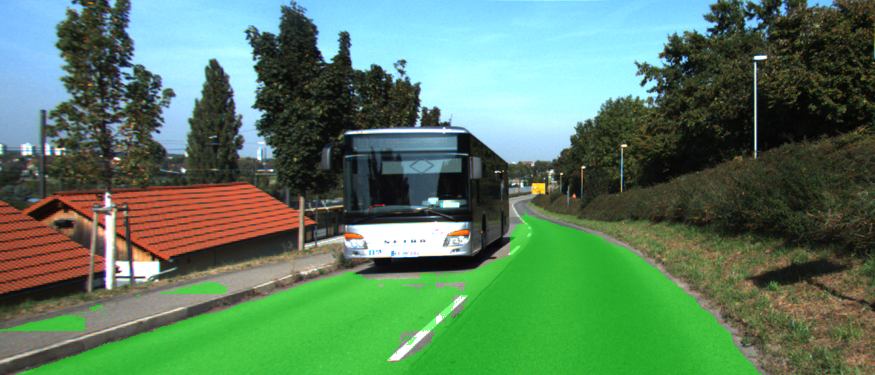}
    \caption{Using regression approach with stride $s=37$}%
\label{fig:reg_stride37}
\end{figure}

\Cref{fig:reg_stride37} shows the output when using a stride of 37. We see that
the edge of the street is classified slightly better. In order to archive this
effect a rather big stride between 37~and~47 is already sufficient. In the
regression approach there is no measurable benefit of using strides below 37.

%% file: ch7-results.tex

\section{Experimental Results}\label{sec:evaluation}

In this section we discuss the experimental results of the models
presented in \cref{sec:model}. First, we evaluate the two different
models. Second, we compare our best one to the work of
Mohan~\cite{Tarel2009} (which is ranked on place~1 at KITTI road
segmentation~\cite{Tarel2009}).
\input{ch3-datasets}

\subsection{Metrics / Experiments}
For the evaluation of the following experiments we used these metrics: accuracy
($\mathbf{ACC}$), average precision ($\mathbf{AP}$), precision
($\mathbf{PRE}$),  recall ($\mathbf{REC}$), false positive rate
($\mathbf{FPR}$),  false negative rate ($\mathbf{FNR}$) and the $F_1$-measure
($\mathbf{F_1}$), see~\crefrange{eq:accuracy}{eq:fMeasure}. Those metrics make
use of true positive (TP), true negative (TN), false positive (FP) and false
negative (FN).

\begin{align}\label{eq:accuracy}
\text{ACC} &= \frac{TP + TN}{TP + FP + TN + FN}\\
\text{PRE} &= \frac{TP}{TP + FP}\label{eq:precision}\\
\text{REC} &= \frac{TP}{TP + FN}\label{eq:recall}\\
\text{FPR} &= \frac{FP}{FP + TN}\label{eq:fpr}\\
\text{FNR} &= \frac{FN}{TP+ FN}\label{eq:fnr}\\
\text{AP} &= \frac{1}{11} \displaystyle\sum_{r \in 0,0.1,\dots 1} \max_{\tilde{r}: \tilde{r} > r} \text{Precision}(\tilde{r})\label{eq:ap}\\
F_1 &= \frac{2 \cdot TP}{2TP +FP +FN}\label{eq:fMeasure}
\end{align}

In \cref{eq:ap}  $\text{Precision}(\tilde{r})$ is the measured precision at
recall $\tilde{r}$~\cite{everingham2010pascal}. These metrics are also used by
the official KITTI evaluation. \\ To be able to evaluate different approaches
we used only the training data of the KITTI data set, as the ground truth for
the test data is not publicly available. We splitted the training data
beforehand 20 to 80 (test/training) in order to be able to measure our
performance. Our best model was submitted for the official KITTI evaluation
(\href{http://www.cvlibs.net/datasets/kitti/eval_road.php}{www.cvlibs.net/datasets/kitti/eval\_road.php})\\

Our goal was to achieve an adequate classification performance while staying
within a time frame of \textbf{\SI{20}{\milli\second}} as maximal
classification time per image. As the normal use case of road segmentation
would be in autonomous cars the real-time ability is a crucial point.\\

We used a computer with these specifications for the experiments (GPU was used
for training and testing):
\begin{itemize}
    \item Intel(R) Core(TM) i7-4790K CPU @ \SI{4.00}{\giga\hertz}
    \item System memory \SI{16}{\gibi\byte}
    \item GeForce GTX 980 \SI{4}{\gibi\byte} RAM
\end{itemize}

\Cref{tab:ownapproach} shows the result of our evaluation and regression
approach using the models and parameters as described in \cref{sec:model}. It
shows clearly that the regression model has an overall better $F_1$-measure and
accuracy score than the classification model. Surprisingly, a smaller stride
does not automatically lead to better performance. The classification model
shows  the best result with a stride of $s=37$, while in the regression based
approach a stride $s=51$ achieves the best performance. Unfortunately the RAM
of the graphic card limited our possibility to use larger strides and patch
sizes. This could have been a promising possibility to train and evaluate on a
full image size and still keep our time constraint and even enhance our
performance.\\

\begin{table}[tbp]
    \centering
    \begin{tabular}{c|cccccc}
        \toprule
        \textbf{Model} & $\mathbf{F_1}$ & \textbf{TN} & \textbf{FP} & \textbf{FN} & \textbf{TP} & \textbf{ACC} \\
        \midrule
        \textbf{Reg., $s=10$} & \SI{88.0}{\percent} & \SI{97.8}{\percent} & \SI{2.2}{\percent}& \SI{19.7}{\percent}& \SI{80.2}{\percent}& \SI{94.7}{\percent}\\
        \textbf{Reg., $s=37$} & \SI{89.0}{\percent}& \SI{97.3}{\percent}& \SI{2.6}{\percent}& \SI{17.6}{\percent}& \SI{82.3}{\percent} &  \textcolor{red}{\SI{94.8}{\percent}}\\
        \textbf{Reg., $s=51$} & \textcolor{red}{\SI{89.5}{\percent}} &\SI{96.9}{\percent} & \SI{3.1}{\percent} & \SI{16.5}{\percent}& \textcolor{red}{\SI{83.5}{\percent}} & \SI{94.6}{\percent}\\
        \midrule
        \textbf{Cla., $s=10$} & \SI{85.4}{\percent} & \SI{98.1}{\percent}& \SI{1.9}{\percent}&\SI{24.1}{\percent} & \SI{75.8}{\percent} & \SI{94.2}{\percent}\\
        \textbf{Cla., $s=37$} & \SI{86.2}{\percent}& \SI{95.9}{\percent} & \SI{4.1}{\percent} & \SI{21.2}{\percent} & \SI{78.7}{\percent} & \SI{92.9}{\percent}\\
        \textbf{Cla., $s=51$} & \SI{70.1}{\percent} & \textcolor{red}{\SI{98.2}{\percent}} & \SI{1.8}{\percent} & \SI{45.1}{\percent} & \SI{54.9}{\percent} & \SI{90.6}{\percent}\\
        \bottomrule
    \end{tabular}
    \caption{Results of classification (cla.) and regression (reg.) models
             with different strides $s$ on our own test set (58~images,
             $ ~6.7 \cdot 10^6$ pixels). The table entries highlighted in
             red are the best evaluation scores regarding different parameterizations.}%
\label{tab:ownapproach}
\end{table}

The predefined time constraint (classification of one image in under
\SI{20}{\milli\second}) was met which is shown by the run time evaluation
displayed in \Cref{tab:runtime}. As expected, the runtime increases with a
smaller stride size. The classification model shows an overall faster run time
performance as the regression. Finally, we meet the time constraint by using a
stride of $s=51$ in both approaches.

\begin{table}[tbp]
    \centering
    \begin{tabular}{c|ccc}
        \toprule
        \textbf{network type / stride $\bm{s}$} & 10 & 37 & 51 \\
        \midrule
        \textbf{regression}     & \SI{1.99}{\second} & \SI{0.29}{\second} & \SI{0.18}{\second} \\
        \textbf{classification} & \SI{1.83}{\second} & \SI{0.20}{\second}  & \SI{0.11}{\second}\\
        \bottomrule
    \end{tabular}
    \caption{Runtime per image ($621 \times 188$ pixel).}%
\label{tab:runtime}
\end{table}

\begin{figure*}[]
    \begin{subfigure}[t]{\columnwidth}
        \includegraphics[width=\columnwidth]{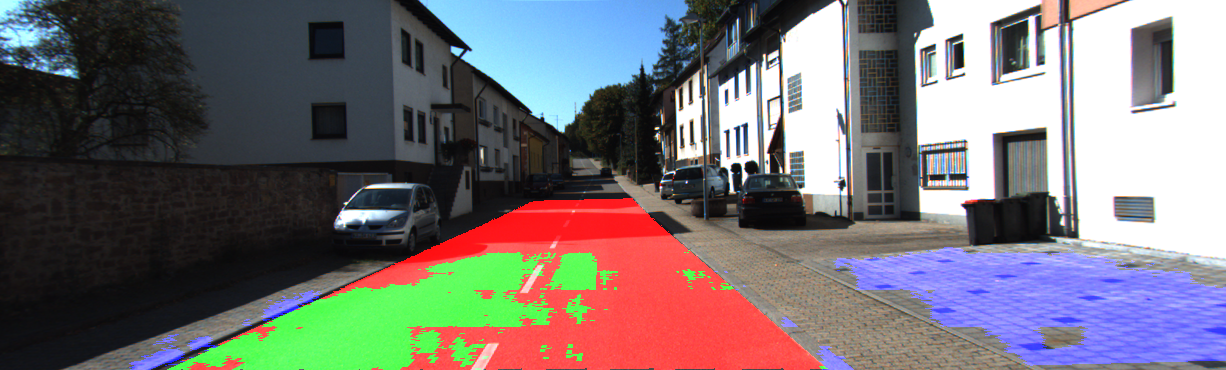}
        \includegraphics[width=\columnwidth]{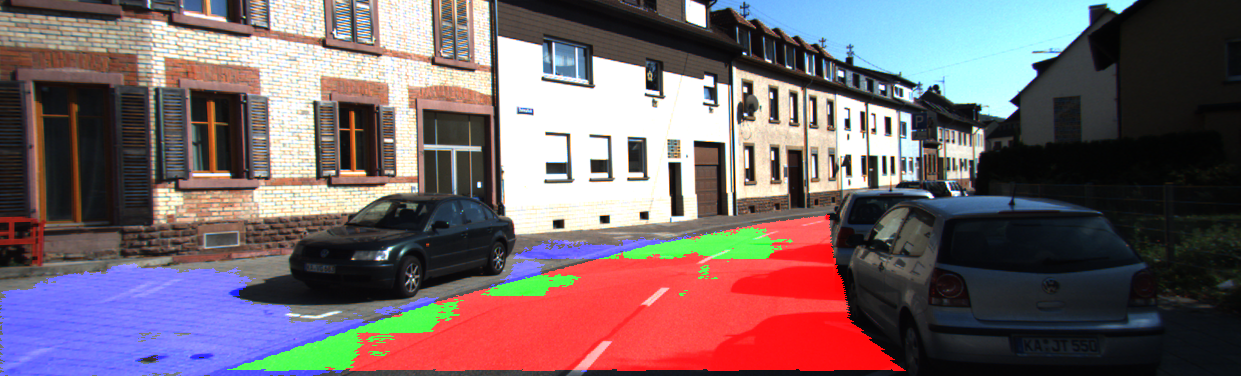}
        \caption{KITTI test data on which our neural net performed badly. Here, red denotes false negatives, blue areas correspond to false positives and green represents true positives.}%
\label{fig:sfig1}
    \end{subfigure}
    \begin{subfigure}[t]{\columnwidth}
        \includegraphics[width=\columnwidth]{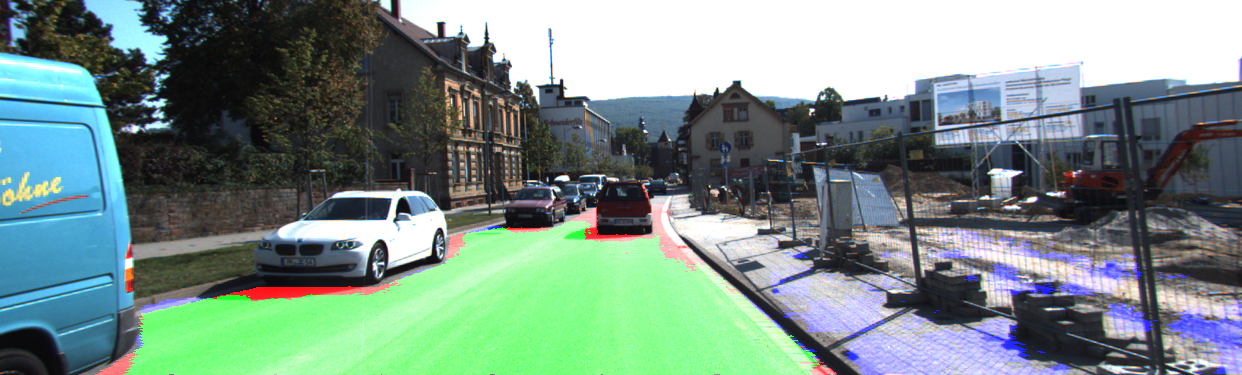}
        \includegraphics[width=\columnwidth]{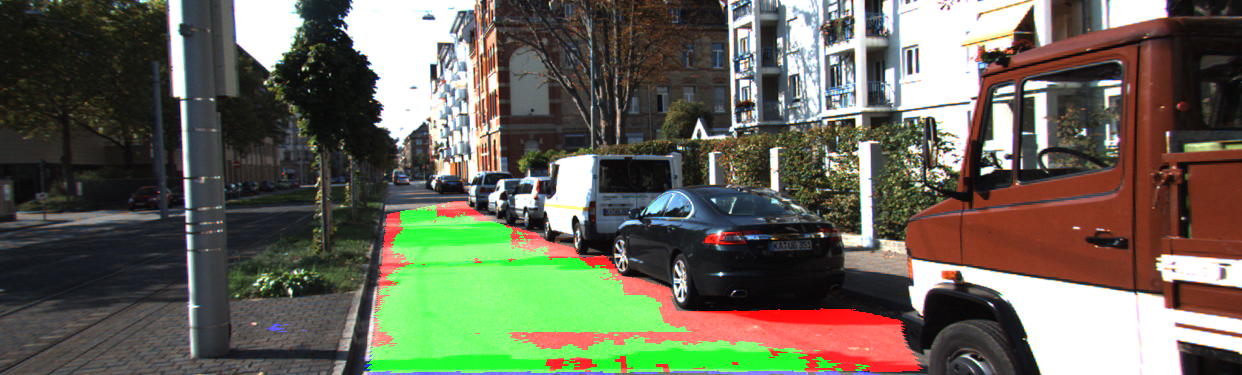}
        \caption{KITTI test data on which our neural net performed well.}%
\label{fig:sfig2}
    \end{subfigure}
\end{figure*}

\begin{table*}[]
    \centering
    \begin{tabular}{c|cccccc}
    \toprule
    \textbf{Benchmark} & $\mathbf{F_1}$ & \textbf{AP} & \textbf{PRE} & \textbf{REC} & \textbf{FPR} & \textbf{FNR}\\
    \midrule
    UM    & \SI{67.91}{\percent} & \SI{61.63}{\percent} & \SI{86.90}{\percent} & \SI{55.74}{\percent} & \SI{3.83}{\percent} & \SI{44.26}{\percent}\\
    UMM   & \SI{79.67}{\percent} & \SI{78.41}{\percent} & \SI{93.29}{\percent} & \SI{69.51}{\percent} & \SI{5.50}{\percent} & \SI{30.49}{\percent}\\
    UU    & \SI{56.48}{\percent} & \SI{51.89}{\percent} & \SI{84.67}{\percent} & \SI{42.37}{\percent} & \SI{2.50}{\percent} & \SI{57.63}{\percent}\\
    URBAN & \SI{71.10}{\percent} & \SI{65.14}{\percent} & \SI{89.83}{\percent} & \SI{58.84}{\percent} & \SI{3.67}{\percent} & \SI{41.16}{\percent}\\
    \bottomrule
    \end{tabular}
    \caption{Results of our regression model (stride 51) on the official KITTI evaluation with different road types.}%
\label{tab:kitti}
\end{table*}

As the regression approach had the best performance and also met our time
constraint, we used it to evaluate the KITTI test set and submitted the results
after a transformation into birds eye view (KITTI
specifications).\Cref{tab:kitti} shows the results which are split into the
different road types (UM, UMM, UU, URBAN). Unfortunately, our regression model
performs much worse on the official test set than on our own test set. Here the
$F_1$-measure score ranges between
\SI{56.4}{\percent} and \SI{79.7}{\percent} while Mohan ~\cite{Tarel2009}
achieves on all road categories a $F_1$-measure score of about
\SI{90.0}{\percent}. The reason for this huge difference might be: \\

\begin{enumerate}
    \item Overfitting of the neural network on our own test data
    \item Specialization of our two models on images with half the original
          size (the KITTI evaluation is done on full size images)
    \item Visible in the two images of \cref{fig:sfig1} is an example of very
          bad performance on a part of the test image data. Basically more
          non-street is classified as street and most of the street is not
          recognized as one at all. This could be due to the fact of shadows
          in some parts of the street and a bit different color of this
          particular street than most of the street our neural network has
          learned in the training.
\end{enumerate}
To improve the latter it would be essential to use training data of a lot more
different street types and lighting conditions.\\
Finally \cref{fig:sfig2} gives some positives example where our neural network
did well. There are hardly no false positives and the street around the cars is
nicely segmented.

%% file: ch3-datasets.tex

\subsection{Data Sets}\label{sec:datasets}
The KITTI Road Estimation data set~\cite{Fritsch2013} was used for training of
the models and for obtaining the experimental results reported in
\cref{sec:evaluation}.

The left color image base kit contains a training and a test set. All photos
are in an urban environment. The training set has 95~photos with land markings
(um), 96~photos with multiple lane markings (umm) and 98~photos where the
street has no lane markings (uu). The test set has 96~um~photos, 94~umm~photos
and 100~uu~photos.

The width of all photos is in $\Set{1226, 1238, 1241, 1242}$, the height is in
$\Set{370, 374, 375, 376}$.

The data photos are given as 8-bit color RGB PNG files. The labels (ground
truth) are given as images of the same size as the data image, but with only
three colors: red (\verb+#ff0000+), magenta (\verb+#ff00ff+) for street and
black (\verb+#000000+) for other streets than the one the car is on.

%% file: ch8-discussion.tex

\section{Conclusion}\label{sec:discussion}

The results presented in this paper were obtained over five months in a
hands-on course. We started with the Caffe framework and experimented with
a fork created by Jonathan Long. The model he provided could not be used on our
computers, because \SI{4}{\gibi\byte} of GPU~RAM were not enough to evaluate
the model. We tried to adjust the model model, but we failed due to the lack of
documentation, cryptic error messages and random crashes while training or
evaluating. This was the reason why we switched to Lasagne. Using this
framework, we noticed that we still need to try many different topologies. Our
first tries lead to bad classification accuracy and we are not aware of any
analytical way to determine a network topology for a given task. For this
reason, we developed the SST framework. This allows developers to quickly
train, test, and evaluate new network topologies. Although the framework got
its final flexible form in the last month of the practical course, we used it
to evaluate regression and classification models with several topologies. The
results are described in \cref{sec:evaluation}.

In standard scenes, the classification accuracy is impressive. The street gets
segmented very well in a runtime of well below  $\SI{0.5}{\second}$. However,
in some images the model does perform very badly. These are mainly images with
special situations such as an uncommon street colors or unusual lightning. We
believe that these problems can be easily eliminated by using more training
data. Another approach to get better results on the KITTI data set is to train
the model with different data and only use the KITTI training data for
fine-tuning.

One advantage of our models is that they are perfectly parallelizable. Each
image section can be evaluated independently. This can be advantageous in
practical applications. When using specialized hardware such as neuromorphic
chips it is possible to build hundreds of cores in a car. In such a case our
classification approach can yield outstanding results. Given enough training
data (e.g. 1.2~million images) using GoogLeNet or AlexNet can provide perfect
classification results.

Finally one can also improve the results with better hardware. For some of our
models the \gls{GPU} RAM was the limiting factor. Especially for the regression
model using a bigger section of the image can lead to much better results in
quality as well as in speed.

%% file: pixel-wise-street-segmentation.bbl
\providecommand{\etalchar}[1]{$^{#1}$}
\begin{thebibliography}{EVGW{\etalchar{+}}10}
\providecommand{\url}[1]{#1}
\csname url@samestyle\endcsname
\providecommand{\newblock}{\relax}
\providecommand{\bibinfo}[2]{#2}
\providecommand{\BIBentrySTDinterwordspacing}{\spaceskip=0pt\relax}
\providecommand{\BIBentryALTinterwordstretchfactor}{4}
\providecommand{\BIBentryALTinterwordspacing}{\spaceskip=\fontdimen2\font plus
\BIBentryALTinterwordstretchfactor\fontdimen3\font minus
  \fontdimen4\font\relax}
\providecommand{\BIBforeignlanguage}[2]{{%
\expandafter\ifx\csname l@#1\endcsname\relax
\typeout{** WARNING: IEEEtranSA.bst: No hyphenation pattern has been}%
\typeout{** loaded for the language `#1'. Using the pattern for}%
\typeout{** the default language instead.}%
\else
\language=\csname l@#1\endcsname
\fi
#2}}
\providecommand{\BIBdecl}{\relax}
\BIBdecl

\bibitem[Aly08]{aly2008real}
M.~Aly, ``Real time detection of lane markers in urban streets,'' in
  \emph{Intelligent Vehicles Symposium, 2008 IEEE}.\hskip 1em plus 0.5em minus
  0.4em\relax IEEE, 2008, pp. 7--12.

\bibitem[BBB{\etalchar{+}}10]{Bergstra2010}
J.~Bergstra, O.~Breuleux, F.~Bastien, P.~Lamblin, R.~Pascanu, G.~Desjardins,
  J.~Turian, D.~Warde-Farley, and Y.~Bengio, ``Theano: a {CPU} and {GPU} math
  expression compiler,'' in \emph{Proceedings of the Python for Scientific
  Computing Conference ({SciPy})}, Jun. 2010, oral Presentation.

\bibitem[BBY90]{Beucher1990}
\BIBentryALTinterwordspacing
S.~Beucher, M.~Bilodeau, and X.~Yu, ``Road segmentation by watershed
  algorithms,'' in \emph{PROMETHEUS Workshop, Sophia Antipolis, France}, 1990.
  [Online]. Available:
  \url{http://citeseerx.ist.psu.edu/viewdoc/summary?doi=10.1.1.22.5731}
\BIBentrySTDinterwordspacing

\bibitem[DSR{\etalchar{+}}15]{sander_dieleman_2015_27878}
\BIBentryALTinterwordspacing
S.~Dieleman, J.~Schlüter, C.~Raffel, E.~Olson, S.~K. Sønderby, D.~Nouri,
  D.~Maturana, M.~Thoma, E.~Battenberg, J.~Kelly, J.~D. Fauw, M.~Heilman,
  diogo149, B.~McFee, H.~Weideman, takacsg84, peterderivaz, Jon, instagibbs,
  D.~K. Rasul, CongLiu, Britefury, and J.~Degrave, ``Lasagne: First release.''
  Aug. 2015. [Online]. Available: \url{http://dx.doi.org/10.5281/zenodo.27878}
\BIBentrySTDinterwordspacing

\bibitem[EVGW{\etalchar{+}}10]{everingham2010pascal}
M.~Everingham, L.~Van~Gool, C.~K. Williams, J.~Winn, and A.~Zisserman, ``The
  pascal visual object classes (voc) challenge,'' \emph{International journal
  of computer vision}, vol.~88, no.~2, pp. 303--338, 2010.

\bibitem[FKG13]{Fritsch2013}
J.~Fritsch, T.~Kuehnl, and A.~Geiger, ``A new performance measure and
  evaluation benchmark for road detection algorithms,'' in \emph{International
  Conference on Intelligent Transportation Systems (ITSC)}, 2013.

\bibitem[GMM12]{6182716}
C.~Guo, S.~Mita, and D.~McAllester, ``Robust road detection and tracking in
  challenging scenarios based on markov random fields with unsupervised
  learning,'' \emph{Intelligent Transportation Systems, IEEE Transactions on},
  vol.~13, no.~3, pp. 1338--1354, Sept 2012.

\bibitem[JSD{\etalchar{+}}14]{Jia2014}
Y.~Jia, E.~Shelhamer, J.~Donahue, S.~Karayev, J.~Long, R.~Girshick,
  S.~Guadarrama, and T.~Darrell, ``Caffe: Convolutional architecture for fast
  feature embedding,'' \emph{arXiv preprint arXiv:1408.5093}, 2014.

\bibitem[KSH12]{krizhevsky2012imagenet}
A.~Krizhevsky, I.~Sutskever, and G.~E. Hinton, ``Imagenet classification with
  deep convolutional neural networks,'' in \emph{Advances in neural information
  processing systems}, 2012, pp. 1097--1105.

\bibitem[LSD14]{long2014fully}
J.~Long, E.~Shelhamer, and T.~Darrell, ``Fully convolutional networks for
  semantic segmentation,'' \emph{arXiv preprint arXiv:1411.4038}, 2014.

\bibitem[Moh14]{mohan2014deep}
R.~Mohan, ``Deep deconvolutional networks for scene parsing,'' \emph{arXiv
  preprint arXiv:1411.4101}, 2014.

\bibitem[PVG{\etalchar{+}}11]{scikit-learn}
F.~Pedregosa, G.~Varoquaux, A.~Gramfort, V.~Michel, B.~Thirion, O.~Grisel,
  M.~Blondel, P.~Prettenhofer, R.~Weiss, V.~Dubourg, J.~Vanderplas, A.~Passos,
  D.~Cournapeau, M.~Brucher, M.~Perrot, and E.~Duchesnay, ``Scikit-learn:
  Machine learning in {P}ython,'' \emph{Journal of Machine Learning Research},
  vol.~12, pp. 2825--2830, 2011.

\bibitem[SLJ{\etalchar{+}}14]{SzegedyLJSRAEVR14}
\BIBentryALTinterwordspacing
C.~Szegedy, W.~Liu, Y.~Jia, P.~Sermanet, S.~Reed, D.~Anguelov, D.~Erhan,
  V.~Vanhoucke, and A.~Rabinovich, ``Going deeper with convolutions,''
  \emph{CoRR}, vol. abs/1409.4842, 2014. [Online]. Available:
  \url{http://arxiv.org/abs/1409.4842}
\BIBentrySTDinterwordspacing

\bibitem[TB09]{Tarel2009}
\BIBentryALTinterwordspacing
J.-P. Tarel and E.~Bigorgne, ``Long-range road detection for off-line scene
  analysis,'' in \emph{Intelligent Vehicles Symposium, 2009 IEEE}, June 2009,
  pp. 15--20. [Online]. Available:
  \url{http://ieeexplore.ieee.org/xpl/login.jsp?tp=&arnumber=5164245&url=http%3A%2F%2Fieeexplore.ieee.org%2Fxpls%2Fabs_all.jsp%3Farnumber%3D5164245}
\BIBentrySTDinterwordspacing

\end{thebibliography}
